\def\year{2015}
\documentclass[letterpaper]{article}
\usepackage{aaai}
\usepackage{times}
\usepackage{helvet}
\usepackage{courier}
\usepackage{amsmath}
\usepackage{amsfonts}
\usepackage{amssymb}
\usepackage{bm}
\usepackage{color}

\usepackage{amsmath}  
\usepackage{graphicx} 
\usepackage{amssymb}
\usepackage{amsthm}

\usepackage{amsmath}
\usepackage{algorithmicx}
\usepackage{algorithm}
\usepackage{algpseudocode}

\theoremstyle{plain}

\theoremstyle{definition}

\theoremstyle{remark}

\newcommand{\bP}{\mathbb P}
\newcommand{\bb}[1]{\textbf{#1}}

\newcommand{\val}{\textit{Val}}

\newcommand{\mc}[1]{\mathcal{#1}}
\newcommand{\VV}{\textit Val}
\newcommand{\bmc}[1]{\bm{\mc {#1}}}
\newcommand{\cl}{\textit{class-}}

\newcommand{\dagg}{}
\newcommand{\dom}{\succcurlyeq}

\makeatletter
\newsavebox\myboxA
\newsavebox\myboxB
\newlength\mylenA

\newcommand*\xoverline[2][0.75]{%
    \sbox{\myboxA}{$\m@th#2$}%
    \setbox\myboxB\null
    \ht\myboxB=\ht\myboxA%
    \dp\myboxB=\dp\myboxA%
    \wd\myboxB=#1\wd\myboxA
    \sbox\myboxB{$\m@th\overline{\copy\myboxB}$}
    \setlength\mylenA{\the\wd\myboxA}
    \addtolength\mylenA{-\the\wd\myboxB}%
    \ifdim\wd\myboxB<\wd\myboxA%
       \rlap{\hskip 0.5\mylenA\usebox\myboxB}{\usebox\myboxA}%
    \else
        \hskip -0.5\mylenA\rlap{\usebox\myboxA}{\hskip 0.5\mylenA\usebox\myboxB}%
    \fi}
\makeatother

\usepackage{color}

\begin{document}
\title{Probabilistic Structural Controllability in Causal Bayesian Networks}
\author{Ardavan S. Nobandegani${}^{\dagger}$ \&  Ioannis N. Psaromiligkos${}^{\dagger}$}

\maketitle
\begin{abstract}
\begin{quote}
Humans routinely confront the following key question which could be viewed as a probabilistic variant of the controllability problem: While faced with an uncertain environment governed by causal structures, how should they practice their autonomy by intervening on driver variables, in order to increase (or decrease) the probability of attaining their desired (or undesired) state for some target variable? In this paper, for the first time, the problem of probabilistic controllability in Causal Bayesian Networks (CBNs) is studied. More specifically, the aim of this paper is two-fold: (i) to introduce and formalize the problem of probabilistic structural controllability in CBNs, and (ii) to identify {a sufficient} set of driver variables for the purpose of probabilistic structural controllability of a generic CBN. {We also elaborate on the nature of minimality the identified set of driver variables satisfies.} In this context, the term ``structural" signifies the condition wherein solely the structure of the CBN is known. 
\end{quote}
\end{abstract}

\section{Introduction} 
The aptitude to perceive causation plays a central role in human cognition. Intervention, as a defining feature of humans to manifest their autonomy, is the sole means of actively (in contrast with the passive mode of being a mere observer) interacting with a world governed by causal structures. Among possible intentions behind exerting intervention, the notion of ``control" is a notable one---that is, informally speaking, to manipulate some variables\footnote{The terms ``node" and ``variable" will be used interchangeably throughout.} (also called driver variables) of a system to, either directly or indirectly, ``shape", ``guide", or ``control" variables of the system which are of interest.

In this work, the problem of Targeted Probabilistic Structural controllability (TPS-controllability) in the context of Causal Bayesian Networks (CBNs) is addressed and formalized. The term `structural' signifies the condition wherein the agent is equipped merely with the causal structure of the domain under study. The term `targeted', on the other hand, emphasizes that the agent is interested in controlling the behavior of a specific subset (or all) of variables in the domain called target variables. Finally, the term `probabilistic' highlights the probabilistic nature of the problem under study.

At a high level, we define the problem of \emph{probabilistic controllability} in the context of CBNs as follows: \emph{How an agent, provided with the knowledge of the set of intervenable variables,  should devise her intervention, i.e., (Q.1) ``which" variables to intervene on, and (Q.2) ``how" to intervene on those, so as to ``control" the behavior of a particular variable(s) of interest in the domain (represented by a CBN), that is, to maximize/minimize the probability of the occurrence of a state of interest for a set of target variables.} 

The problem of probabilistic \emph{structural} controllability in the context of CBNs is then accordingly defined as that of probabilistic controllability---as stated above---with one crucial additional constraint on the agent's part: The agent is solely equipped with the knowledge of the underlying causal \emph{structure} of the domain (i.e., the CBN's topology) and is uninformed of the parametrization thereof. In this work, we aim at identifying the minimal set of intervenable variables sufficient for TPS-controllability of an \emph{arbitrary} CBN. Particularly, we devise an algorithm, $\mc C^\ast$, which identifies a sufficient set of intervenable variables for the purpose of TPS-controllability of a generic CBN. We also elaborate on the nature of minimality the identified set satisfies. Furthermore, to formalize the problem under study and to articulate the results,  along the way: (i) we devise, building upon Pearl's concept of \emph{stochastic policy} (cf. \cite{pearl2000causality}, Sec. 4.2), a graphical representation of a generic intervention policy, and (ii) put forth a hierarchical construct for intervention policies wherein moving up in the hierarchy amounts to empowering the agent to exercise more advanced forms of intervention.

The question of interest to this work has significant ramifications for studies on strategic planning and policy making in domains enjoying causal structure. Equally importantly, the problem under study has notable connections to how humans, at the computational level of analysis \cite{marr1982vision} and in line with the rational analysis approach \cite{anderson1990adaptive}, devise their intervention strategies efficiently to increase the odds of attaining their desired goals while faced with their uncertain environment.

\section{Notation and Terminology}
In this section, we present some preliminary notations and terminologies which will be adopted in this paper. Random quantities are denoted by bold-faced letters; their realizations are denoted by the same letter but non-bold. More specifically, Random Variables (RVs) are denoted by lower-case bold-faced letters, e.g., $\bb{x}$, and their realizations by non-bold lower-case letters, e.g., $x$. Likewise, sets of RVs are denoted by bold-faced calligraphic letters, e.g., $\bmc X$, and their corresponding realizations by non-bold calligraphic letters, e.g., $\mc X$. $\textit{Val}(\cdot)$ denotes the set of possible values a random quantity can take on. To simplify presentation, we incur the following abuse of notation: We denote the  probability $\bP(\bb{x}=x)$ by $\bP(x)$ for some RV $\bb{x}$ and its realization $x\in \textit{Val}(\bb{x})$. For conditional probabilities, we will use the notation $\bP(x|y)$ instead of $\bP(\bb{x}={x}|\bb{y}={y})$. Likewise, $\bP(\mc X|\mc Y):=\bP(\bmc X=\mc X|\bmc Y=\mc Y)$ for $\mc X \in \VV(\bmc X)$ and $\mc Y \in \VV(\bmc Y)$. Random quantities are assumed to be discrete, throughout, unless stated otherwise.

Throughout the paper, the Directed Acyclic Graph (DAG) $G=(V,E)$ characterizes the non-intervened causal structure of the domain where $V$ denotes the set of nodes/variables and $E$ denotes the set of edges. We adopt Pearl's notation $do(x):=do(\bb x=x)$ to denote an atomic intervention on $\bb x$ so as to force it to take on value $x$. Also, $ip(\bb x)$ denotes the intervention policy to be adopted for $\bb x$ the meaning of which will be clarified in the subsequent section; informally intervention policy refers to how the agent decides to manipulate the intervened variable (cf. \cite{pearl2000causality}, Sec. 4.2). Intervention policy may or may not functionally depend on other variables of the domain. As we will see later, intervention policy in its most generic form is nothing but a Conditional Probability Distribution (CPD). $\delta(\cdot)$ denotes the Kronecker delta function. Also, Backward Chaining (BC) on a variable refers to the simple process of identifying its parents (i.e., immediate causes) and the parents of the parents and so forth until the boundaries of the CBN are reached.
 
Before proceeding further, let us formally define two key notions, namely, \emph{subsumability} and \emph{domination}.

\textbf{Def. (Subsumability):}
DAG $\dagg G_1=(V_1,E_1)$ \emph{subsumes} DAG $\dagg G_2=(V_2,E_2)$, denoted in short by $\dagg G_1 \supseteq \dagg G_2$, iff $V_1=V_2$ and $E_2 \subseteq E_1$. We refer to the set $E_1\setminus E_2$, as the \emph{surplus} of $\dagg G_1$ with respect to $\dagg G_2$. 

\textbf{Def. (Domination):}
DAG $\dagg G_1=(V_1,E_1)$ \emph{dominates} DAG $\dagg G_2=(V_2,E_2)$, denoted by $\dagg G_1 \succcurlyeq \dagg G_2$, iff there \emph{exists} a parametrization of $\dagg G_1$ which yields a result for the objective of interest that is no worse than the best achievable by \emph{any} possible parametrizations of $\dagg G_2$. For instance, if the objective of interest is to maximize the probability of some event of interest, say $\bb r=r$ for some $r\in \VV(\bb r)$, then we write $\dagg G_1 \succcurlyeq \dagg G_2$ iff there exists a parametrization of $\dagg G_1$ which yields some value for the probability of interest, $\bP(r)$, which is greater than or equal to the best achievable by any possible parametrizations of $\dagg G_2$.

\textbf{Lemma 1. (Domination vs Subsumability):} Let $G_1, G_2$ be DAGs. Then,  $\dagg G_1 \supseteq \dagg G_2 \Rightarrow \dagg G_1 \succcurlyeq \dagg G_2  $.

\textbf{Proof:} The proof is straightforward once we realize that one can very well take advantage of the extra edges of $\dagg G_1$ with respect to $\dagg G_2$ (i.e., the surplus of $\dagg G_1$ with respect to $\dagg G_2$) which give one more ``degrees of freedom" to entertain and hence to achieve a result which is equally good or better than the best achievable by all possible parametrizations of $\dagg G_2$ in terms of the objective of interest.\hfill $\blacksquare$

In subsequent sections where we introduce potential objectives of interest, the above statements become clearer.

\section{Motivating Example}
\label{Sec_motivating_example}
To develop some intuition as to the problem under study, a series of informative examples will be presented.

\begin{figure}[h!]
\centering
\includegraphics[width=0.25\textwidth]{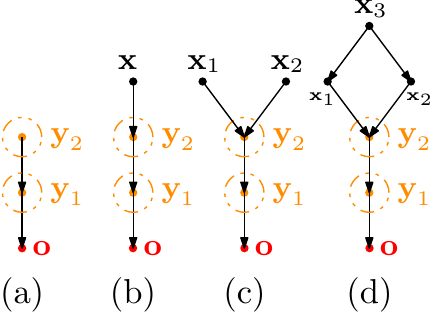}
\caption{Motivating example.}
\label{fig_motive_1}
\end{figure}

Let us first consider the CBN depicted in Fig. \ref{fig_motive_1}(a). Variables $\bb y_1, \bb y_2$ are amenable to intervention (or, in short, \emph{intervenable}). Let us assume that the objective is to make the occurrence of the event $\bb o=o \in \VV(\bb o)$ as likely as possible through intervening on any subset of variables $\{\bb y_1, \bb y_2\}$ (or to choose not to intervene at all corresponding to choosing the empty set). The key question is how, by mere investigation of the \emph{structure} of the CBN depicted in Fig. \ref{fig_motive_1}(a), to decide: (i) on which (intervenable) variable(s) to intervene, and (ii) how the intervention should be exercised (a notion referred to as \emph{Intervention Policy} (IP)). It is easy to come to the conclusion that, to make the occurrence of $\bb o=o$ as likely as possible, one needs to just intervene on $\bb y_1$ and force it into the state $\bb y_1=y_1^\ast$ where $y_1^\ast$ is the realization for $\bb y_1$ conditioned on which the probability of event $\bb o=o$ is maximum, i.e., $y_1^\ast=\arg \max_{y_1 \in \val(\bb y_1)} \bP(o|y_1)$. 

It is crucial to realize then that, due to the \emph{structure} of the CBN depicted in Fig. \ref{fig_motive_1}(a), and \emph{regardless} of its parametrization, it suffices for the agent to solely intervene on $\bb y_1$ for the purpose of TPS-controllability of $\bb o=o$ (the answer to (i)). Furthermore, since the agent is assumed to be equipped merely with the structure of the underlying CBN and not the paramterization thereof, based on the above argument on $\bb y_1$'s IP, the agent can just arrive at the conclusion that $\bb y_1$'s IP has the functional (or structural) form of\footnote{In fact, the agent can reason out one step further and come to the conclusion that $\bb y_1$'s IP must have the functional form of $\bP(\bb y_1)=\delta(\bb y_1=y^\ast)$, however, the agent cannot identify/specify the value of $y^\ast$---due to the lack of knowledge about the paramtrization of the CBN.} $\bP(\bb y_1)$---that is, merely the \emph{non-parametric} form of the IP (the answer to (ii)). {Altogether, a solution to the problem of TPS-controllability of $\bb o=o$ is $\{\bb y_1\}$ (which is a sufficient set of variables to be intervened) along with $\bP(\bb y_1)$ which is the \emph{functional} form of $\bb y_1$'s IP.} Following the same line of reasoning for the CBNs depicted in Figs. 1(b-d), it is straightforward to argue that $\{\bb y_1\}$ is a sufficient set for TPS-controllability of the target variable $\bb o$, and $\bb y_1$'s IP has the functional form of $\bP(\bb y_1)$ akin to what we had for Fig. \ref{fig_motive_1}(a).

\begin{figure}[h!]
\centering
\includegraphics[width=0.09\textwidth]{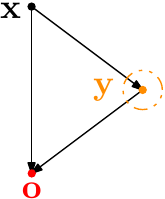}
\caption{Motivating example.}
\label{fig_motive_2}
\end{figure}

Let us consider another example that highlights a key idea, namely, that we need to broaden our understanding of the notion of intervention (cf. \cite{pearl2000causality}, Sec. 4.2). Consider the CBN depicted in Fig. \ref{fig_motive_2}. This time, only $\bb y$ is intervenable. Assume that (only for this particular example), all the variables are binary-valued; the prior probability on $\bb x$ is $\bP(\bb x)$ (which is assumed to be non-degenerate), $\bb y:=\neg \bb x$, and $\bb o :=\bb x \oplus \bb y$ where $\neg$ and $\oplus$ denote the logical connectives \emph{not} and \emph{xor}, respectively. It is easy to verify that the event $\bb o=1$ occurs with probability one, regardless of the choice of $\bP(\bb x)$. For the problem of TPS-controllability of $\bb o=1$, the agent has to decide whether or not to intervene on $\bb y$. Imagine an intervention were to be exercised on $\bb y$. Whether the agent would set $\bb y=0$, or $\bb y=1$, the objective event of $\bb o=1$ would become less likely to happen compared to that of the (non-intervened) original model. At first glance, the fact that exercising intervention makes the situation worse in terms of the objective of interest seems rather counter-intuitive. How could it be that having the freedom to manipulate variable $\bb y$ (which even happens to be one of the parents of the objective node) whatever way we like does not allow you to outperform the (non-intervened) original model? The answer lies in developing a better understanding of the term ``whatever way we like." For that purpose, we need to broaden our conception of the notion of intervention and go beyond practicing merely a primitive atomic form of intervention denoted by $do(\bb y=y)$ in the literature. A more advanced form of intervention is to pick the state to which we want to force the intervened variable as a function of the states of some other variables of the domain. That is, IP may depend functionally on a collection of other variables in the domain. In this example, choosing $\bb y$'s IP, denoted by $ip(\bb y)$, to functionally depend on (the state of) variable $\bb x$ ensures that, the outcome achieved by exerting such intervention, is equally good or better (no worse than) than that of the (non-intervened) original model. In such a setting, we simply adopt the following terminology/notation: The intervention pair $(\bb y, ip(\bb y):=\bP(\bb y|\bb x))$ (comprising, in order, of the set of intervened variables and their corresponding IPs) is equally good or better than both: (i) the intervention pair $(\bb y,ip(\bb y):=\bP(\bb y)=\delta(\bb y=y))$ corresponding to the simplistic atomic form of intervention on $\bb y$ discussed above which does not depend on the state of $\bb x$ and, likewise, (ii) the intervention pair $(\varnothing,\varnothing)$ corresponding to the original model (without intervention). A simple comparison between the CBN in Fig. \ref{fig_motive_1}(a) and the one in Fig. \ref{fig_motive_2} reveals the following: Had there been left, upon intervening on $\bb y$, no paths through which $\bb x$'s causal effect on the objective variable $\bb o$ could be mediated, there would be no need for the intervention policy of $\bb y$ to functionally depend on $\bb x$ (akin to Fig. \ref{fig_motive_1}(a)). 

\section{Intervention Policy}
\label{intervention_gen}
The notion of IP delineates how an intervened variable should be ``manipulated." More specifically, IP indicates whether other variables play any role or not (and if so, how) in devising how the manipulation on a to-be-intervened variable is to be practiced. It is intuitive that the more variables we are allowed to functionally depend on while devising the IP of a to-be-intervened variable, the more ``degrees of freedom" we have in controlling the behavior of the to-be-intervened variable. Next, we will formalize this intuition by, first, putting forward a simple graphical representation of intervention policy and, then, by using the idea of \emph{subsumability} which draws on the fact that the original non-intervened model could be thought of as a special case of the intervened model. By allowing a larger number of variables for the IP of an intervened variable to \emph{functionally} depend on, we show that IPs can be organized in a hierarchical construct wherein moving up in the hierarchy amounts to empowering the agent to exercise more sophisticated forms of intervention. 

\subsection{Hierarchical Construct}
\label{intervention_gen_hierarchy}
Before proceeding further let us make a definition: The scope of an IP is the set of variables, except the intervened variable itself, that the IP functionally depends on. For instance, for $ip(\bb y):=\bP(\bb y|\bb s)$, the scope is comprised of the variable $\bb s$. In short, the idea of organizing IPs into a hierarchical construct is inspired by the simple realization that, by delimiting the set of variables the agent is allowed to incorporate into the scope of the intervened variable's IP (functionally represented by a conditional probability distribution), we impose a constraint on the expressive power of the intervention policy. 

The following notation henceforth will be employed to refer to different IP-classes:

$\bullet$ IP \cl$0$: This class refers to the set of IPs where the scope of each is the empty set, i.e., a setting wherein the IP(s) of the intervened variable(s) is not allowed to incorporate any variables into its scope. That is, if variable $\bb x$ happens to be decided to be intervened and the agent is only permitted to adopt \cl$0$ intervention policies, then, the agent is just allowed to place IP of the functional form $\bP(\bb x)$ on $\bb x$ to exercise her intervention. It is crucial to note that, the agent is allowed to parameterize $\bP(\bb x)$ as she wishes, yet the functional form of the IP is constrained. 

$\bullet$ IP \cl$1$: This class refers to the set of IPs where the scope of each comprises the immediate causes of the corresponding intervened variable. That is, in this class, the IP of the intervened variable $\bb x$ has the functional form $ip(\bb x):=\bP(\bb x|par(\bb x))$ where $par(\bb x)$ is the set of immediate causes for $\bb x$. That is, the scope of \cl$1$ IP $ip(\bb x)$ is $par(\bb x)$. 

$\bullet$ IP \cl$j, \forall j\geq 2$: This class refers to the set of IPs where the scope of each is the ancestors of the corresponding intervened variable up to $i^\text{th}$ level. For instance, for the case of IP \cl$2$, IP(s) of the intervened variable(s) is solely allowed to take into account the state of (i) the immediate causes, and (2) the immediate causes of the variables in (i), thereby, altogether functionally depending on all the ancestors up to the $2^\text{nd}$ level. 

$\bullet$ IP \cl$\infty$: This class refers to the set of IPs where the scope of each is \emph{all} the ancestors of the corresponding intervened variable. Note that the complete set of ancestors of a variable $\bb x$ can be found by instantiating the BC on $\bb x$.

Finally, it is crucial to notice the following. For an IP to be in a particular class amounts to imposing a constraint solely on the \emph{functional} form of the IP; the agent is free to choose any parametrization for the IP as she may wish. Therefore, $ip(\bb x)\in \cl i$ simply means that the \emph{functional} form of $ip(\bb x)$ is constrained in accord to the definition of IP \cl$i$ given above, yet, it could be arbitrarily parameterized. Also, assuming that $\bmc X=\{\bb x_i\}_{i=1}^m$, the notation $ip(\bmc X)\in \cl j$ will be adopted as a shorthand for the following: $ip(\bb x_i)\in \cl j, \forall i=1,\cdots,m.$  

\subsection{Graphical Representation}
\label{intervention_gen_graphical_rep}
In this section, we develop a way of visualizing IPs. The idea is simple. If the IP of a to-be-intervened variable $\bb x$ happens to functionally depend on $\bb y$, then we show this by a directed dash-dotted arrow emanating from $\bb y$ and pointing towards $\bb x$. To ensure that any practice of intervention is fully expressed by such edges we introduce the following convention: For DAG $G=(V,E)$, a clamped variable $\mathfrak{C}$ is added\footnote{It is implicitly assumed throughout this paper that the variable $\mathfrak{C}$ has been added to DAG $G$ \emph{a priori}. Also, we will not depict $\mathfrak{C}$ in the figures unless needed.} to $V$. Then, intervening on a variable $\bb a$ which has no parents and exerting $ip(\bb a)=\bP(\bb a)$ will be illustrated graphically by a dash-dotted edge emanating from $\mathfrak{C}$ towards $\bb a$.  In general, upon $\bb y$ taking on the state $y$, the agent may decide to set the value of $\bb x$ to a fixed value $x$ (deterministic IP), or to set the value of $\bb x$ probabilistically (stochastic IPs), i.e., $\bb x$ {takes on} values from $\VV(\bb x)$ according to some conditional probability distribution $\bP(\bb x|\bb y)$. In both cases, $ip(\bb x)$ is said to \emph{functionally} depend on variable $\bb y$. Simply put, in devising the intervention policy of $\bb x$, namely, $ip(\bb x)$, the state of $\bb y$ is taken into account. The notion of probabilistic IP is discussed in (\cite{pearl2000causality}, pp. 113-114) under the title of \emph{stochastic policy}.

\begin{figure}[h!]
\centering
\includegraphics[width=0.19\textwidth]{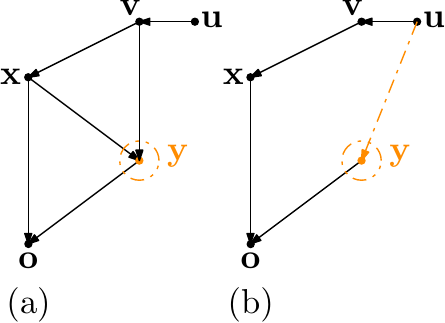}
\caption{Sample case. \textbf{(a):} Original CBN. Variable $\bb y$ is to be intervened according to $ip(\bb y):=\bP(\bb y|\bb u)$. \textbf{(b):} The graphical representation of intervening on $\bb y$ with $ip(\bb y):=\bP(\bb y|\bb u)$. The figure simply illustrates the fact that the state of $\bb y$ gets decided (potentially probabilistically) according to the state of $\bb u$. Notice that according to \cite{pearl2000causality}, upon intervening on $\bb y$, all the incoming edges into $\bb y$ should be removed.}
\label{fig_graph_rep_int}
\end{figure}

Let us first give some definitions which will prove useful in the subsequent sections. For the given definitions, DAG $\dagg G=(V,E)$ represents the causal structure of the domain.
  
\textbf{Def. (Intervention Pair):}
A set of intervened variables $\bmc K\subseteq V$ along with their corresponding IPs comprise a pair, called an \emph{intervention pair}, which is denoted by $(\bmc K, ip(\bmc K))$.

\textbf{Def. (Intervention DAG (\emph{i}-DAG)):}
Every intervention pair $(\bmc K, ip(\bmc K))$ for $\bmc K\subseteq V$ uniquely specifies a DAG (as described above) which we refer to as the \emph{i}-DAG associated to that intervention pair. The \emph{i}-DAG associated to $(\bmc K, ip(\bmc K))$ is denoted by $(\bmc K, ip(\bmc K))_{\dagg G}$.

\textbf{Def. (\emph{i}-Subsumability):} 
For $\bmc X,\bmc Y \subseteq V$, \emph{i}-DAG $(\bmc X,ip(\bmc X))_G$ \emph{i}-subsumes \emph{i}-DAG $(\bmc Y,ip(\bmc Y))_G$, denoted in short by $(\bmc X,ip(\bmc X))_G \supseteq_i (\bmc Y,ip(\bmc Y))_G$, iff (i) $(\bmc X,ip(\bmc X))_G \supseteq (\bmc Y,ip(\bmc Y))_G$, (ii) the set of the dash-dotted edges in $(\bmc Y,ip(\bmc Y))_G$ is a subset of the set of the dash-dotted edges in $(\bmc X,ip(\bmc X))_G$, and (iii) the surplus of $(\bmc X,ip(\bmc X))_G$ with respect to $(\bmc Y,ip(\bmc Y))_G$ is solely comprised of dash-dotted edges.

\textbf{Def. (\emph{i}-Domination):} 
For $\bmc X,\bmc Y \subseteq V$, \emph{i}-DAG $(\bmc X,ip(\bmc X))_G$ \emph{i}-dominates \emph{i}-DAG $(\bmc Y,ip(\bmc Y))_G$, denoted by $(\bmc X,ip(\bmc X))_G \dom_{i} (\bmc Y,ip(\bmc Y))_G$ for short, iff there exist a parameterization for the dash-dotted edges (see Fig. \ref{fig_graph_rep_int}) in $(\bmc X,ip(\bmc X))_G$ which yields a result for the objective of interest that is no worse than the best achievable by any possible parametrizations of the dash-dotted edges in \emph{i}-DAG $(\bmc Y,ip(\bmc Y))_G$.

\textbf{Lemma 2. (\emph{i}-Domination vs \emph{i}-Subsumability):} \emph{
Let DAG $\dagg G=(V,E)$ characterize the causal structure of the domain. Let $(\bmc X,ip(\bmc X))_G$ and $(\bmc Y,ip(\bmc Y))_G$ be two \emph{i}-DAGs for some $\bmc X,\bmc Y \subseteq V$. Then, the following holds:} 
\begin{eqnarray*}
\small (\bmc X,ip(\bmc X))_G \supseteq_i (\bmc Y,ip(\bmc Y))_G \Rightarrow (\bmc X,ip(\bmc X))_G \dom_i (\bmc Y,ip(\bmc Y))_G.
\end{eqnarray*}

\textbf{Proof:} The rationale is similar to the one presented for Lemma 1.\hfill $\blacksquare$

It immediately follows (due to Lemma 2) that, for $\dagg G=(V,E)$ and $\forall \bmc K\subseteq V$, the following holds true: $\forall j\geq m$, $(\bmc K,ip(\bmc K) \in \cl j)_{\dagg G}\dom_i (\bmc K,ip(\bmc K) \in \cl m)_{\dagg G}.$

\section{TPS-Controllability of CBNs: Formalization}
\label{sec_P_con}
Let DAG $\dagg G=(V,E)$ characterize the causal structure of the domain. Let $V=V_i\cup \bar{V}_i$ where $V_i$ denotes the set of nodes/variables amenable to intervention (or, in short, \emph{intervenable}), and $\bar{V}_i$ be the complement of $V_i$, i.e., $\bar{V}_i=V\setminus V_i$. The probability of interest, in its generic form takes the form\footnote{The connection to Pearl's notation for $do$-calculus is as follows (cf. \cite{pearl2000causality}, p. 114): $\bP(y)|_{{\bP}^\ast(x|z)} = \bP(y|do[\bb x;{\bP}^\ast(x|z)])$.} $\bP(\bmc O=\mc O|do[\bmc X;ip(\bmc X)=ip(\mc X)])$, or in short $\bP(\mc O|do[\bmc X;ip(\mc X)])$, where $\bmc O$ denotes the set of target variables, $\mc O$ denotes the realization of interest, $\bmc X$ denotes the set of intervened variables, $ip(\bmc X)$ denotes the \emph{functional} form (i.e., non-parametric representation) of the to-be-adopted intervention policy, and $ip(\mc X)$ denotes a specific parametrization\footnote{Which is equivalent to a specific parametrization of the dash-dotted edges representing the intervention policy exercised on variables $\bmc X$ in the corresponding $i$-DAG (see Fig. \ref{fig_graph_rep_int}).} of $ip(\bmc X)$. Also, $do[\bmc X;ip(\bmc X)]$ denotes the setting wherein variables $\bmc X$ are intervened according to $ip(\bmc X)$ (functional form) and, likewise, $do[\bmc X;ip(\mc X)]$ denotes the setting wherein variables $\bmc X$ are intervened according to $ip(\bmc X)=ip(\mc X)$. One can write, $\bmc O=\bmc O_i \cup \bar{\bmc O}_i$ where $\bmc O_i \subseteq V_i$ and $\bar{\bmc O}_i \subseteq \bar{V}_i$. Objectives of interest could have any of the following forms:
\begin{eqnarray}
\min_{\bmc X \subseteq V_i}\bigg(\min_{ip(\mc X)\in \cl \infty}\bP(\mc O|do[\bmc X;ip(\mc X)])  \bigg),\label{min-min}\\
\max_{\bmc X \subseteq V_i}\bigg(\max_{ip(\mc X)\in \cl \infty}\bP(\mc O|do[\bmc X;ip(\mc X)])  \bigg),\label{max-max}
\end{eqnarray}
and,
\begin{eqnarray}
\min_{\bmc X \subseteq V_i}\bigg(\max_{ip(\mc X)\in \cl \infty}\bP(\mc O|do[\bmc X;ip(\mc X)])  \bigg),\label{min-max}\\
\max_{\bmc X \subseteq V_i}\bigg(\min_{ip(\mc X)\in \cl \infty}\bP(\mc O|do[\bmc X;ip(\mc X)])  \bigg).\label{max-min}
\end{eqnarray}

In the sequel, we focus on objectives (\ref{min-min}) and (\ref{max-max}). The discussion on objectives (\ref{min-max}) and (\ref{max-min}) is deferred to the end of the paper. Therefore, whenever we use the statement ``objective of interest" we are specifically referring to both of objectives (\ref{min-min}) and (\ref{max-max}) unless stated otherwise.

Next, we devise an algorithm, $\mc C^{\ast}$, for the problem of TPS-controllability of CBNs. $\mc C^{\ast}$ outputs a set of intervenable variables, $\bmc X^\ast$, which is ``optimal" with respect to objectives (\ref{min-min}) and (\ref{max-max}). In other words, $\bmc X^\ast$ is a sufficient choice of variables to intervene on (according to IP \cl$\infty$) to satisfy objectives (\ref{min-min}) and (\ref{max-max}). To put it formally, in the next section we show that for $\bmc X^\ast$ the following holds:  $\forall \bmc Y \subseteq V_i$, 
\begin{eqnarray*}
\label{eq_optimality}
(\bmc X^\ast,ip(\bmc X^\ast)\in\cl\infty)_{\dagg G}\dom_i (\bmc Y,ip(\bmc Y)\in\cl\infty)_{\dagg G}. 
\end{eqnarray*} 

\section{TPS-Controllability of CBNs: Algorithm $\mc C^{\ast}$}
\label{Sec_alg_c_star}
\begin{figure}[h!]
\centering
\includegraphics[width=0.16\textwidth]{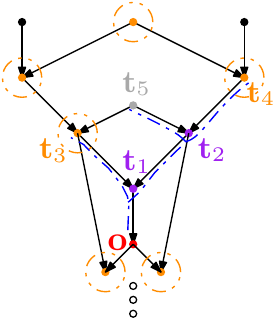}
\caption{Sample Case: Variable $\bb o$ (depicted in red) is the target variable. The intervenable variables (i.e., members of $V_i$) are circled. {The BC execution paths are colored in blue and illustrated by dash-dotted lines.} Upon initiating the BC at the target variable $\bb o$, we arrive at the $\bb t_1$ (depicted in purple) located at the junction. Next, we arrive at $\bb t_2$ and $\bb t_3$. Since $\bb t_3 \in V_i$, the BC terminates at $\bb t_3$. On the other hand, since $\bb t_2\not\in V_i$, the BC continues. Having performed the BC on $\bb t_2$, we arrive at $\bb t_4$ and $\bb t_5$. Since $\bb t_4 \in V_i$, the BC terminates on $\bb t_4$. At the end, since $\bb t_5$ (depicted in grey) has no parents (immediate causes), the BC terminates at $\bb t_5$ as well. Therefore, by mere investigation of the structure, $\mc C^\ast$ outputs the set $\bmc X^\ast=\{\bb t_3,\bb t_4\}$ as a solution to the objectives (\ref{max-max}) and (\ref{min-min}) for this particular setting (i.e., the given CBN and the corresponding $V_i$ and $\bb o$ as the target variable).}
\label{fig_alg_c_star}
\end{figure}

Let us explain simply how $\mc C^{\ast}$ works. BC has to be initiated on nodes in $\bmc O$. Upon reaching any node in $ V_i$, the BC execution path terminates at that node. This procedure continues until, for all of the BC execution paths, either: (i) The BC execution path gets terminated at some node belonging to $V_i$, or (ii) a node with no parents is reached. The set of intervenable variables at which BC terminates constitute $\mc C^{\ast}$'s output denoted by $\bmc X^\ast$. Fig. \ref{fig_alg_c_star} depicts a sample execution of $\mc C^{\ast}$.

\subsection{Algorithm $\mc C^{\ast}$: Justification}
\label{sec_c_star_proof}
\textbf{On the Sufficiency of $\bmc X^\ast$:} 
Let us now present the sufficiency proof for $\bmc X^\ast$ where by sufficiency we mean that:  intervening (according to IP \cl$\infty$) on any variables in addition to $\bmc X^\ast$ does not yield any improvement upon what is achievable through merely intervening (according to IP \cl$\infty$) on $\bmc X^\ast$. 

Notice that we can write $V_i=V_i^{BC}\cup \bar{V}_i^{BC}$ where $V_i^{BC}$ is the set of intervenable variables which belong to the subgraph generated by executing BC on the target nodes $\bmc O$. It is obvious that intervening on any variable in $\bar{V}_i^{BC}$ is pointless due to the following argument\footnote{This statement immediately follows from Rule 3 of Pearl's $do$-calculus (cf. \cite{pearl2000causality}, p. 95).}: ($\star$) variables in $\bar{V}_i^{BC}$ have no direct or indirect causal effect on any of the target nodes. Now, what is left to be shown is why, among all $V_i^{BC}$, it suffices to intervene on $\bmc X^\ast$ (according to IP \cl$\infty$) and, why, intervening on any additional variables does not improve upon what is achievable through intervening merely on $\bmc X^\ast$. More formally, the question is why the following holds: $\forall \bmc Y \subseteq V_i^{BC}$
\begin{eqnarray*}
(\bmc X^\ast,ip(\bmc X^\ast)\in\cl\infty)_{\dagg G}\dom_i (\bmc Y,ip(\bmc Y)\in\cl\infty)_{\dagg G}.
\end{eqnarray*}
Notice that, the extension to the case of $\forall \bmc Y \subseteq V_i$ immediately follows from argument ($\star$). The realization of the fact that variable $\bb y \in V_i^{BC}$ was not selected (for intervention) by $\mc C^{\ast}$ implies that $\bb y$'s causal effect on the target variables which are descendant\footnote{Intervening on $\bb y$, obviously, could merely influence $\bb y$'s descendant.} of $\bb y$ must have been mediated through some of the selected nodes by $\mc C^{\ast}$ say $\bmc Y^{\dagger} \subseteq \bmc X^{\ast}$ (otherwise, $\bb y$ would have been selected). The claim as to the redundancy of further intervening on $\bb y$ in addition to exerting intervention (according to IP \cl$\infty$) on $\bmc Y^{\dagger}$ is as follows. 
Let $G=(V,E)$ be the DAG associated to the (non-intervened) underlying causal structure of the domain. First, notice that the effect of stochastic policies can be expressed in terms of atomic interventions in $G$ as explained in (\cite{pearl2000causality}, pp. 113-114) and (\cite{pearl1995causal}, p. 684).  Due to Rule 2 of Pearl's $do$-calculus, intervention on $\bb y$ can be exchanged with $\bb y$ being merely passively observed. Subsequently, due to Rule 1 of Pearl's $do$-calculus, $\bb y$'s observation can be dismissed. Notice that intervening on $\bmc Y^{\dagger}$ according to IP \cl$\infty$ (which amounts to incorporating all the ancestors of $\bmc Y^{\dagger}$ into their IPs) renders $\bb y$ $d$-separated from the target variables which are descendant of\footnote{Incorporation of all the antecedents of $\bmc Y^{\dagger}$ into their IPs guarantees that all the back-door paths from $\bb y$ to the target variables which are descendant of $\bb y$ are blocked (cf. \cite{pearl2000causality}, Sec. 3.3.1).} $\bb y$, hence the applicability of Rule 2 and Rule 1. This concludes the proof.

\section{On the Minimality of $\bmc X^\ast$}
Let us present two definitions and a proposition which bears on the minimality of $\bmc X^\ast$.

\textbf{Def. \small{(Locally Structurally Minimal (LSM)}\normalsize):}
$\mc C^\ast$'s output, $\bmc X^\ast$, is LSM with respect to DAG $G$ iff there exists a parametrization of $G$ such that \emph{no} proper subset of $\bmc X^\ast$, namely, $\bmc X^{\ast\ast}$, exists for which the following holds:
\begin{eqnarray*}
(\bmc X^{\ast\ast},ip(\bmc X^{\ast\ast})\in \cl\infty)_{\dagg G} \dom_i (\bmc X^{\ast}, ip(\bmc X^{\ast})\in \cl\infty)_{\dagg G}.
\end{eqnarray*}

\textbf{Def. \small{(Uniformly Structurally Minimal (USM)}\normalsize):}
$\mc C^\ast$'s output, $\bmc X^\ast$, is USM iff $\bmc X^\ast$ is LSM with respect to any DAG $G$.

\textbf{Proposition 1.} \emph{$\bmc X^\ast$ is USM with respect to objective (1).}

\textbf{Proof:} The proof is constructive. The objective of interest is maximax given in (1). Let us assume, without loss of generality, that all the RVs are binary-valued and the desired state is for all the target variables to take on value one. Our goal is to parameterize an \emph{arbitrary} $G$ in such a way that: (i) the desired state happens with probability one if variables $\bmc X^{\ast}$ are all set to one through exerting atomic interventions, and (ii) the desired state happens with probability zero otherwise. Start at $\bmc X^{\ast}$. Parameterize the CPD of each $\bb x^\ast \in \bmc X^\ast$ such that it always takes on the value zero. Moving along the BC execution paths terminated at $\bmc X^{\ast}$, proceed towards the target variables which are descendants\footnote{For any target variable $\bb q$ which is not a descendant of $\bmc X^{\ast}$, parameterize $\bP(\bb q|par(\bb q))$ such that $\bb q$ takes the value 1 with probability one.} of $\bmc X^{\ast}$. Along the way, parameterize the CPD associated to any variable $\bb k$ such that, conditioned on $\bb k$'s parents which are descendants of $\bmc X^{\ast}$ (denoted by $par_{\bmc X^\ast}(\bb k)$), $\bb k$ takes on value one iff all $par_{\bmc X^\ast}(\bb k)$ take value one\footnote{In other words, the CPD of $\bP(\bb k|par(\bb k))$ is paremeterized in such a manner that the parents of $\bb k$ which are \emph{not} descendants of $\bmc X^\ast$ are rendered ineffective.}. In other words, intermediate variables like $\bb k$ work as an \emph{and} logical gate. Proceed in the aforementioned manner until all the target variables (which are descendants of $\bmc X^{\ast}$) are reached. It is easy to verify that indeed the desired state happens with probability one iff all the variables $\bmc X^{\ast}$ are set to one and, furthermore, intervening on any proper subset of $\bmc X^{\ast}$ in any way does not yield such an outcome.  This concludes the proof. \hfill $\blacksquare$

\section{On Minimax/Maximin Objectives}
\label{sec_minimax_maximin}
In this section we claim that, subject to the constraint that the IP's of the to-be-intervened variables has to belong to IP \cl$j$, the solution to both minimax and maximin problem is the empty set, for all $j\geq 1$. Let us present a lemma using which it is easy to justify the claim made above.

\textbf{Lemma 3.}
\emph{Let DAG $\dagg G=(V,E)$ represent the causal structure of the domain. $\forall j\geq 1$ and $\forall \bmc X \subseteq V_i$, the following inequalities hold:
\begin{eqnarray*}
\min_{\substack{ip(\mc X)\in \cl j}}\bP(\mc O|do[\bmc X;ip(\mc X)])\leq \bP(\mc O),\\
\bP(\mc O)\leq \max_{\substack{ip(\mc X)\in \cl j}}\bP(\mc O|do[\bmc X;ip(\mc X)]).
\end{eqnarray*}}
The proof for Lemma 3 is straightforward due to the simple realization that, $\forall j\geq 1$, $(\bmc X, ip(\bmc X)\in \cl j)_{\dagg G}$ \emph{i}-subsumes the original CBN, thus $(\bmc X, ip(\bmc X)\in \cl j)_{\dagg G} \dom_i (\varnothing, \varnothing)_{\dagg G}$, $\forall j\geq 1$. Using Lemma 3, it is easy to justify the claim we made earlier as to the solution to the minimax and maximin problems; the solution to both is the empty set. For details, the reader is referred to Appendix. 

\section{Related Work}
In this section, we give an overview of the ideas explored in the literature which are, in spirit, related to the problem under study in this work. The idea of Structure Control Theory (SCT) proposed by Lin \cite{lin1974structural} in the context of Linear Time-Invariant (LTI) systems governed by first-order differential equations (a.k.a. state equations) perhaps comes closest to our problem. In such domains, all variables are deterministic and the states of variables change in time according to the dynamics represented by state equations.

Authors in \cite{liu2011controllability}, drawing on the idea of SCT proposed by Lin \cite{lin1974structural}, aim at identifying the minimal set of variables which are sufficient for the purpose of structural controllability of a generic large-scale LTI system. In a subsequent work, authors in \cite{gao2014target}, relax the objective of structural controllability of the system in whole, to merely that of a particular set of desired variables called target variables. This line of thought has been motivated due to the understanding that in large-scale systems, it may neither be attainable nor required to control the full system but, rather, to merely control a subset of the variables of the system (analogous to target variables in our problem) which are deemed pivotal for the realization of the task at hand. In this light, \cite{gao2014target} is concerned with the very same question underlying our work, yet, perusing it in radically different settings. In \cite{liu2011controllability,gao2014target} both variables and their inter-connections are deterministic in nature whereas, in our case, both have probabilistic natures; a point of departure which leads to a substantially different line of work---both semantically and syntactically. 

\section{Conclusion}
In this paper, for the first time, the problem of TPS-controllability in the context of CBNs was investigated and formalized. Algorithm $\mc C^\ast$ was devised to identify a sufficient set of intervenable variables for the purpose of TPS-controllability of a generic CBN; the minimality of $\mc C^\ast$'s output was also characterized. The provided results can have significant ramifications for studies on strategic planning and policy making allowing one to efficiently practice her interventions in order to maximize (minimize) the odds of the desired (undesired) outcomes.

\newpage
\pagebreak
\bibliographystyle{aaai}
\bibliography{ref}
\newpage
\pagebreak

\section*{Appendix}

\subsubsection{On Minimax/Maximin Objectives: Details}
Using Lemma 3 in the paper, the next inequalities follows. $\forall j\geq 1$,
\begin{eqnarray*}
\max_{\bmc X\subseteq V_i} \bigg(\min_{\substack{ip(\mc X)\in \cl j}}\bP(\mc O|do[\bmc X;ip(\mc X)])\bigg) \leq \bP(\mc O),\\
\bP(\mc O) \leq \min_{\bmc X\subseteq V_i} \bigg(\max_{\substack{ip(\mc X)\in \cl j}}\bP(\mc O|do[\bmc X;ip(\mc X)])\bigg).
\end{eqnarray*}
One can readily conclude from the above inequalities the claim made in the paper as to the solution to the minimax and maximin problems; the solution to both is the empty set.

\subsubsection*{Minimality of $\bmc X^\ast$ for Minimin Objective}
We present the result in a form of a proposition.

\textbf{Proposition A.1:}
\emph{Let DAG $G=(V,E)$ characterize the causal structure of the domain and $\bmc O_i \neq \varnothing$. Any set $\{\bb o^\ast\}$ where $\bb o^\ast \in \bmc O_i$ is a solution to the minimin objective given in (2).}

\textbf{Proof:} Setting $\bb o^\ast \in \bmc O_i$ to any value other than the desired one simply renders the desired event $\bmc O=\mc O$ impossible which subsequently yields value zero for the minimin objective. This concludes the proof. \hfill $\blacksquare$ 
\\
\begin{large}
\begin{center}
\textbf{On the Connection to Psychology}
\end{center}
\end{large}
Algorithm $\mc C^{\ast}$ captures one very common-sensical strategy an individual (referred to as the agent throughout in the paper) follows while faced with the task under study: ``The closer a to-be-intervened node is to the target node(s), the better." That is, there exists this  tendency to intervene on the nodes which are closest\footnote{The extreme case being to intervene right on the target node(s) and set it to the desired value.} to the target node(s). We refer to this intuitive tendency as the \emph{proximity principle}. In psychology literature \cite{glymour2003learning,krynski2007role}, CBN is adopted as a normative model, at the computational level of analysis \cite{marr1982vision}, to represent and reason about causal domains. Algorithm $\mc C^{\ast}$, therefore, could be viewed as a potential proposal at Marr's algorithmic level of analysis  \cite{marr1982vision}, for the problem of how subjects devise their intervention strategies to ``control" the state of a target node. The plausibility of $\mc C^{\ast}$ as a potential algorithmic-level candidate is elevated due to $\mc C^{\ast}$'s capturing nicely, by its mere design (i.e., starting at target nodes and moving backwards toward intervenable nodes), the common-sensical proximity principle.
\end{document}